\pdfoutput=1
\documentclass[11pt]{article}

\usepackage[]{acl}

\usepackage{times}
\usepackage{latexsym}
\usepackage{graphicx}

\usepackage[T1]{fontenc}
\usepackage[utf8]{inputenc}

\usepackage{microtype}

\usepackage{amsfonts}
\usepackage{amsmath}
\usepackage{booktabs}
\usepackage{multirow}
\usepackage{subcaption}

\usepackage{graphicx}
\usepackage{algorithm}
\usepackage{algpseudocode}
\algrenewcommand\algorithmicrequire{\textbf{Input:}}
\algrenewcommand\algorithmicensure{\textbf{Output:}}
\usepackage{caption}

\title{Investigating Information Inconsistency in Multilingual Open-domain Question Answering}

\author{Shramay Palta, ~~Haozhe An, ~~Yifan Yang, ~~Shuaiyi Huang, ~~Maharshi Gor\\
  Department of Computer Science \\
  University of Maryland, College Park \\
  \texttt{\{spalta, haozhe, yang7832, huangshy, mgor\}@umd.edu} \\
  }

\begin{document}
\maketitle
\begin{abstract}
Retrieval based open-domain QA systems use retrieved documents and answer-span selection over retrieved documents to find best-answer candidates. %
We hypothesize that multilingual Question Answering (QA) systems are prone to information inconsistency when it comes to documents written in different languages, because these documents tend to provide a model with varying information about the same topic. 
To understand the effects of the biased availability of information and cultural influence, we analyze the behavior of multilingual open-domain question answering models with a focus on retrieval bias. 
We analyze if different retriever models present different passages given the same question in different languages on TyDi QA and XOR-TyDi QA, two multilingual QA datasets. 
We speculate that the content differences in documents across languages might reflect cultural divergences and/or social biases.
\end{abstract}

\section{Introduction}
The answer to a question is not always unique under different cultural or language backgrounds. 
When one asks a search engine\footnote{\url{https://www2.bing.com/}} in English, "How old do you need to be to be able to drink?", the first relevant passage retrieved states "nationally, the legal drinking age is 21." However, if the same question is asked in Chinese, the search engine retrieves a Chinese passage that indicates, "Usually one has to be at least 16 years old to drink, but there is no minimum legal age." Given the same question in different languages, it is possible that a QA system gives inconsistent answers as a result of training on a corpus that reflects various cultural, historical, and social landscapes.
We hypothesize that information bias in documents causes models to deliver varying performance. Information bias occurs when key information is either measured, collected, or interpreted differently in the training documents. We empirically investigate the validity of our hypothesis by comparing and contrasting the performance of QA models trained on multilingual dataset.

In addition, in-depth investigations of cultural and/or social biases, specifically in the context of open-domain question answering, are still under explored. 
We speculate that the content differences in documents across languages are effective indicators of some cultural divergences and/or social biases. 

In this work, the focus is on understanding retrieval inconsistencies in multilingual open-domain question answering. Specifically, given a question in any language (e.g. Korean), we feed it into a multilingual passage retriever to retrieve multilingual passages as its context from a huge Wikipedia corpus covering 13 languages including English, Arabic, Finnish, Japanese, Korean, Russian, Bengali, Telugu, Indonesian, Thai, Hebrew, Swedish, and Spanish. We then propose a strategy to get paired questions (questions having the same content but in different languages) and their in-language context. In this way, the difference in retrieved context given questions in different languages (e.g. Korean context for a Korean question, English context for a paired English question) would be a good indicator of information inconsistency in a multilingual setting. 

In order to have a quantitative analysis of retrieval differences, we feed the question and its retrieved in-language passage into a shared multilingual answer generator and directly compare the final answer results. We investigate two different settings (oracle and non-oracle setting) for a comprehensive analysis.

\begin{figure*}[t]
	\centering
    \includegraphics[width=\linewidth]{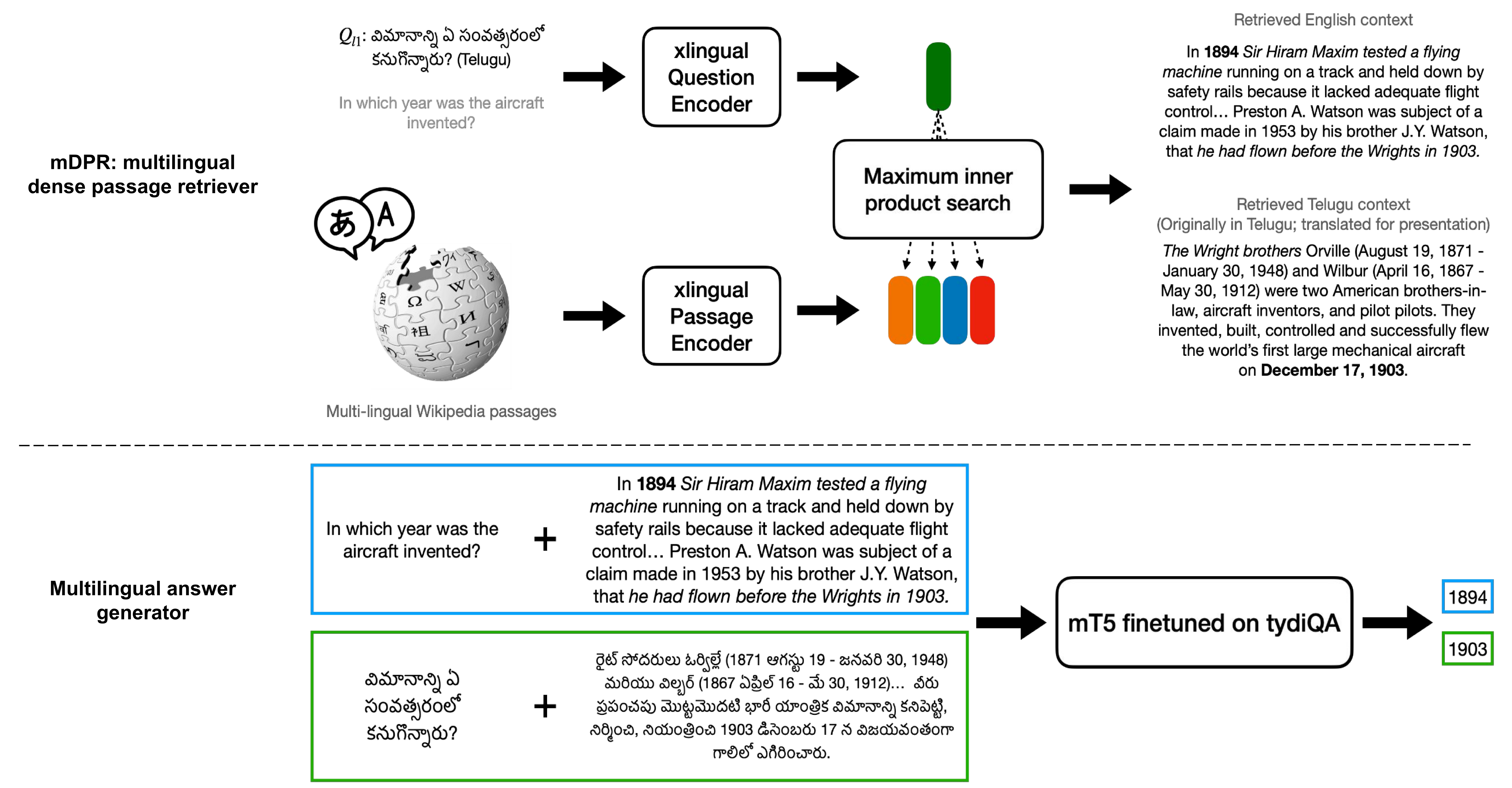}
	\caption{Our pipeline consists of a multilingual dense passage retriever and a multilingual answer generator. Given a question in any language, it is first fed into the question encoder to get an encoding of the question. Relevant multilingual Wikipedia passages are retrieved via maximum inner product search (MIPS). The question and its retrieved in-language context are fed into multilingual answer generator to generate the final answer.}
	\label{fig:pipeline}
\end{figure*}

Our contributions are summarized as follows:
\begin{itemize}
    \item We investigate whether QA datasets exhibit information discrepancy or not.
    \item We propose a strategy to obtain paired questions for cross-lingual analysis.
    \item We evaluate to what extent each component in the QA system contributes to inconsistent answers on the XOR-TyDi QA benchmark across different languages, with a focus on retrieval inconsistencies.
\end{itemize}

\section{Methods}

\subsection{Overview}
An overview of our pipeline is shown in Fig.~\ref{fig:pipeline}. Our framework consists of a multilingual Dense Passage Retriever (mDPR) and a multilingual answer generator. Given a question in any language, we feed it into the question encoder to get an encoding of the question. We retrieve relevant multilingual passages from a huge Wikipedia corpus covering multiple languages by comparing the similarity between question encodings and passage encodings. We select the proper in-language context as its final retrieved context. We then feed the question and its retrieved context into the multilingual answer generator to generate the final answer. We will first describe the details of the mDPR in Sec.~\ref{sec:mDPR} followed by the multilingual answer generator in Sec.~\ref{sec:mag}. Finally, we elaborate on the two settings for analysis as described in Sec.~\ref{sec:analysis}.

\subsection{Multilingual Dense Passage Retriever}
\label{sec:mDPR}
We follow \citet{asai2020xor} for training a multilingual dense passage retriever.\footnote{\url{https://github.com/mia-workshop/MIA-Shared-Task-2022}} We use a multilingual dense passage retriever fine-tuned on XOR-TyDi QA benchmark. The details are as below:

\paragraph{Training:}
mDPR extends Dense Passage Retriever (DPR;~\citet{karpukhin2020dense}) to a multilingual setting. mDPR uses an iterative training approach to fine-tune a pre-trained multilingual language model (e.g., mBERT;~\citet{devlin2018bert}) to encode passages and questions separately. Once training is done, the representations for all passages are computed offline and stored locally. We refer the reader to \cite{asai2020xor} for more details. 

\paragraph{Inference:}
We get a passage encoding from multilingual BERT given a passage with fixed-length sequence of tokens from multilingual documents. At inference, mDPR independently obtains a d-dimensional (d = 768) encoding of the questions from multilingual BERT. It retrieves \textit{k} passages with the \textit{k} highest relevance scores to the question, where the relevance score between a passage and a question \textit{q} is estimated by the inner product of their encoding vectors.

With the trained DPR, we retrieve relevant passages across all available languages including English, Arabic, Finnish, Japanese, Korean, Russian, Bengali, Telugu, Indonesian, Thai, Hebrew, Swedish, and Spanish. For each question in its original language, we get its paired question by translating it into other languages and applying the mDPR module for retrieval.

\subsection{Multilingual Answer Generator}
\label{sec:mag}
We follow ~\citet{clark2020tydi} for training the multilingual answer generator.\footnote{\url{https://github.com/google-research-datasets/tydiqa}} Our multilingual answer generator is MT5 ~\cite{xue-etal-2021-mt5} pretrained from m4C corpus covering 101 languages. We then finetune the MT5 answer generator on the TyDi-QA training split~\cite{clark2020tydi} covering 11 languages.

\subsection{Inconsistency and Error Analysis}
\label{sec:analysis}
\paragraph{Oracle Setting:}
For the oracle setting, we use the ground truth answer as a filter to select the context and the questions to be fed into the answer extractor for analysis. Specifically, we keep the top $20$ retrieval results. We keep questions whose retrieval results contain context in at least two languages, and have the ground truth answers contained within the multilingual languages.

\paragraph{Non-Oracle Setting:}
For the non-oracle setting, we do not use the ground truth answer as a filter to select the context and the questions. Instead,  we only keep the \textit{top-one} score passage. We believe that using the oracle gives bias towards retrieving consistent answers across different languages, and by removing the filter, we can acquire samples that are more likely to reveal information inconsistency.

\begin{figure*}[t]
	\centering
	\begin{subfigure}[]{0.48\linewidth}
        \includegraphics[width=\linewidth]{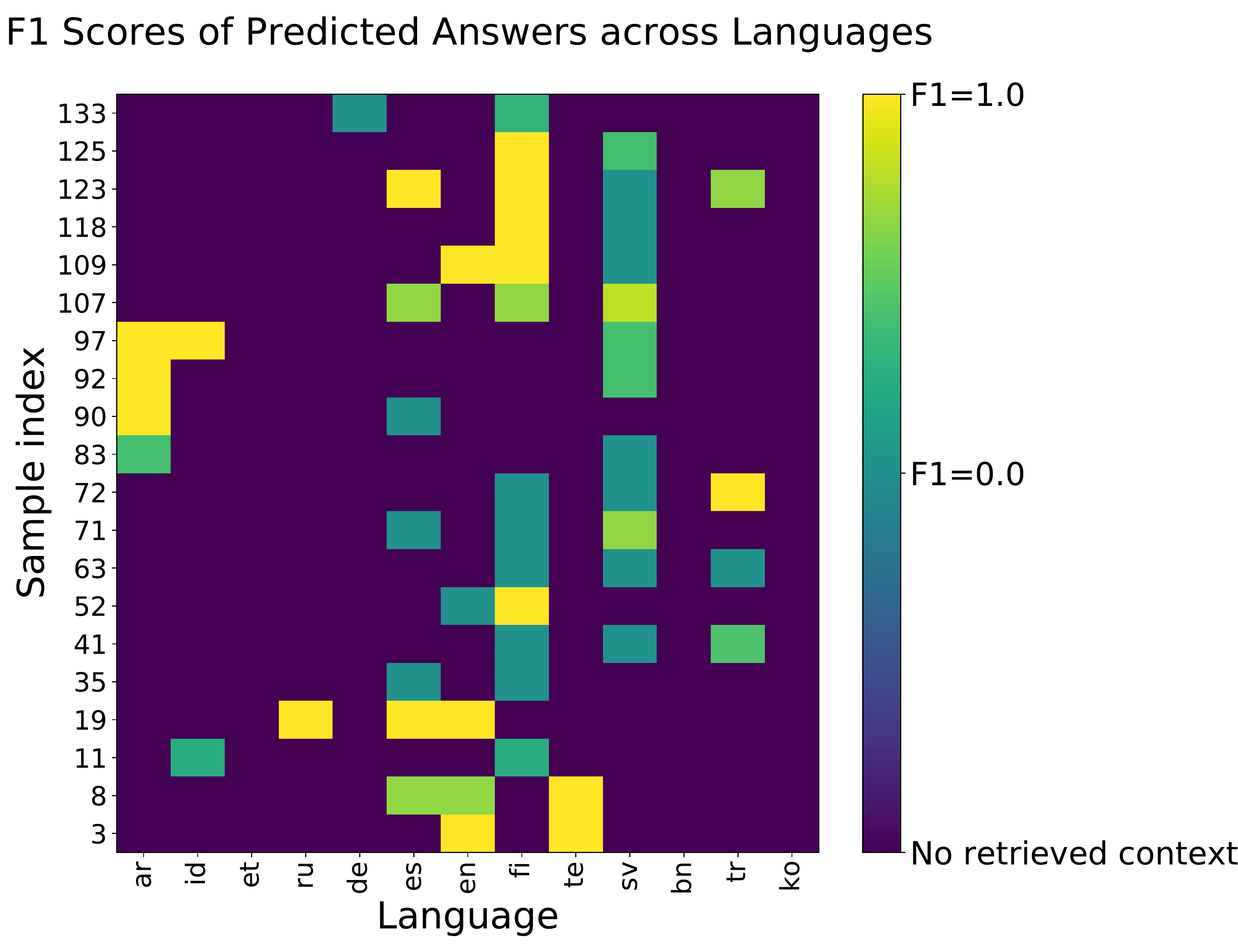}
        \caption{}
        \label{fig:inconsist_dist_gt_filter}
    \end{subfigure}
	\hfill
	\begin{subfigure}[]{0.48\linewidth}
		\centering
		\includegraphics[width=\linewidth]{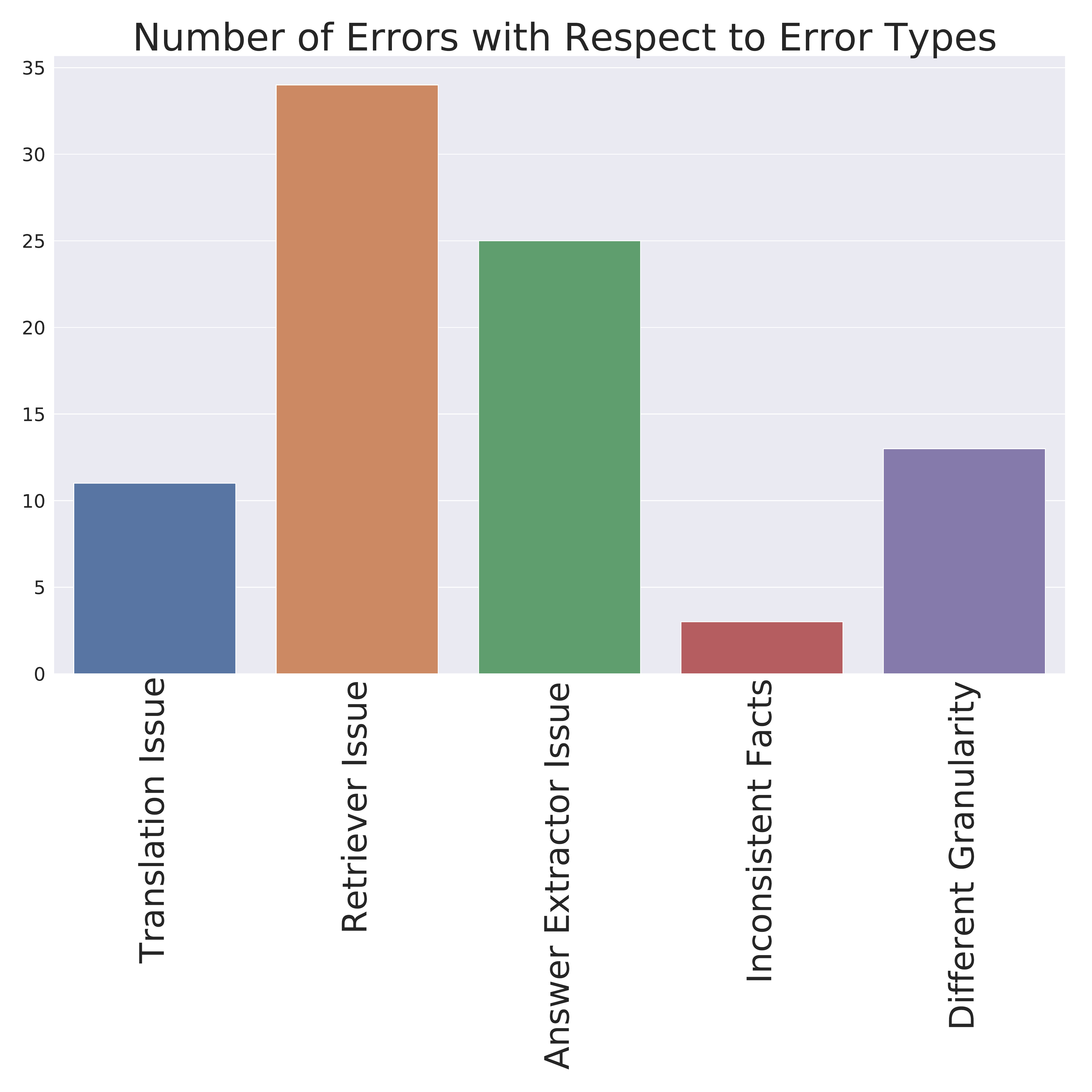}
		\caption{}
		\label{fig:err_type_gt_filter}
	\end{subfigure}
	\caption{Analysis in Oracle Setting: (a) Inconsistent answer distribution across multiple languages. Each row stands for one question with the same meaning but translated into different languages. Given a selected row, the different color indicates the different F1 scores obtained from questions asked in different languages. (b) Error type distribution across languages. Refer text for details.}
	\label{fig:gt_filter_dist}
\end{figure*}

\begin{figure*}[t]
	\centering
	\begin{subfigure}[]{0.45\linewidth}
        \includegraphics[width=\linewidth]{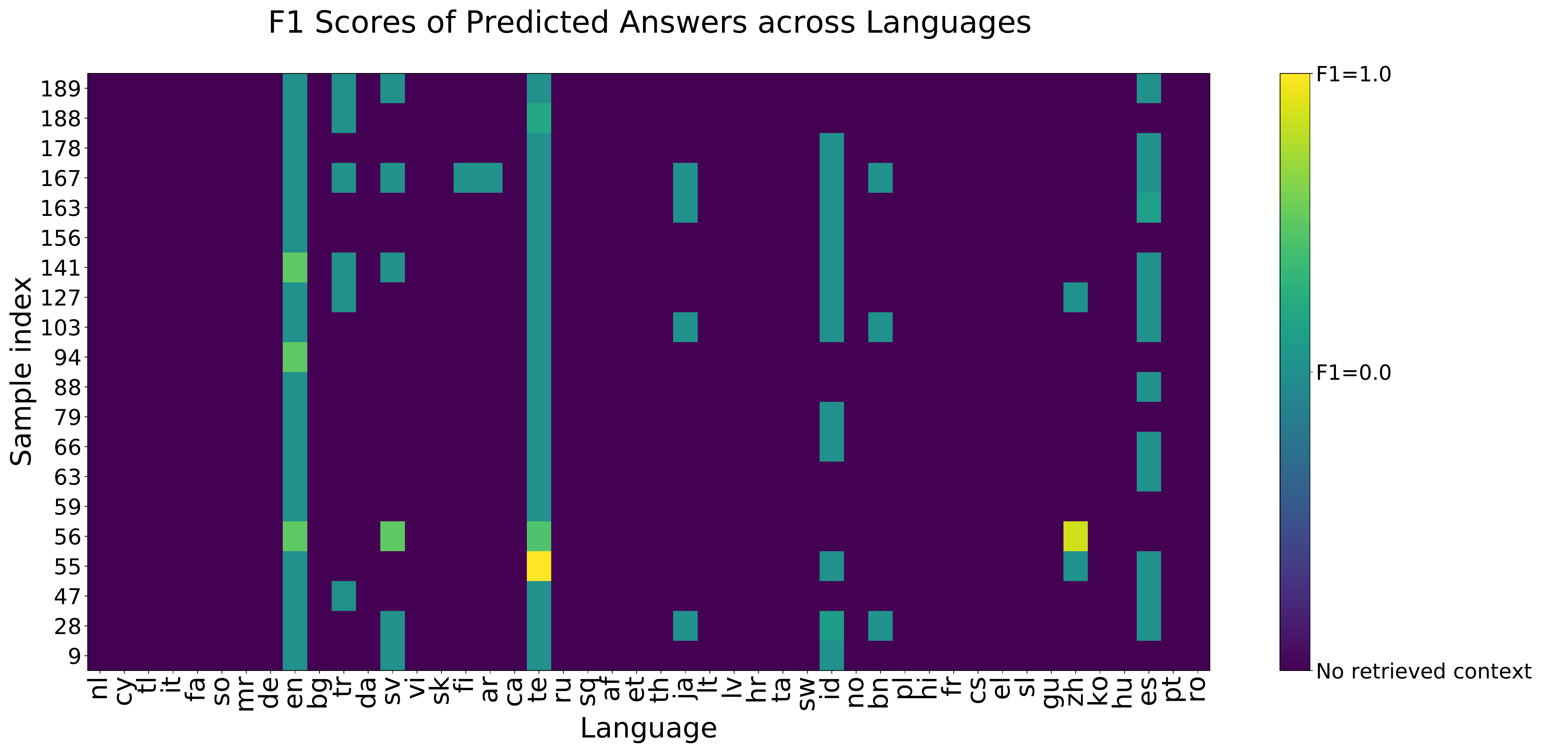}
        \caption{}
        \label{fig:inconsist_dist_nogt_filter}
    \end{subfigure}
	\hfill
	\begin{subfigure}[]{0.45\linewidth}
        \includegraphics[width=\linewidth]{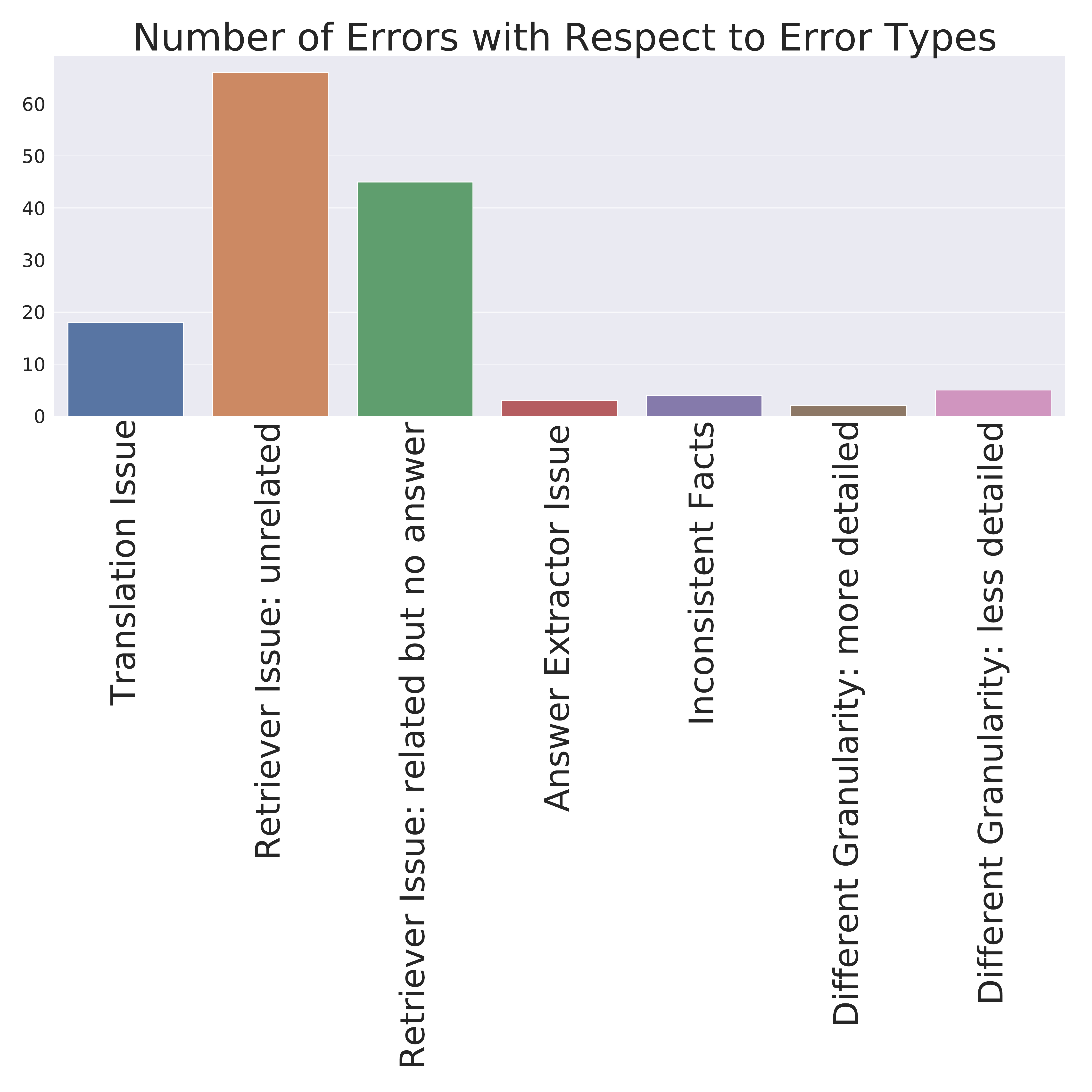}
        \caption{}
      	\label{fig:err_type_nogt_filter}
    \end{subfigure}
	\caption{Analysis in Non-Oracle Setting: (a) Inconsistent answer distribution across multiple languages. Each row stands for one question with the same meaning but translated into different languages. Given a selected row, the different color indicates the different F1 scores obtained from questions asked in different languages. (b) Error type distribution across languages.}
	\label{fig:nogt_filter_dist}
\end{figure*}

\section{Experiments}
In this section, we first elaborate on our implementation details in Sec.~\ref{subsec:detail}, and follow that with the quantitative and qualitative results in Sec.~\ref{subsec:result}.
\subsection{Implementation Details}
\label{subsec:detail}
\subsubsection{Datasets}

\textit{TyDi QA}~\cite{clark2020tydi} is a question answering dataset containing 204K question-answer pairs in 11 typologically diverse languages. It contains instances of language phenomena that would not be found in English-only corpora. In order to reduce the priming effects and provide a realistic and a robust information seeking task, the questions in TyDi QA are written by people who don't know the answer, and want to know what it is, (unlike SQuAD \cite{rajpurkar2016squad} and its descendants) and the data is collected directly in each language without implementing any translation module (unlike MLQA \cite{lewis2019mlqa} and XQuAD \cite{Artetxe:etal:2019}). We use TyDi QA for fine-tuning our multilingual answer extractor.

\textit{XOR-TyDi QA} is a dataset based on TyDi QA.  To support research in multilingual open-domain QA, ~\citet{asai2020xor} constructed a dataset, XOR-TyDi QA, of 40K annotated questions across 7 diverse non-English languages. The questions in this dataset are inherited from TyDi QA (Clark et al., 2020) and consists of questions that TyDi QA could not find the same-language answers for. The rationale behind choosing XOR-TyDi QA is that it is the first large-scale cross-lingual open-retrieval QA dataset that consists of information-seeking questions from native speakers and multilingual reference documents. We use the XOR-TyDi QA training split for fine-tuning our multilingual Dense Passage Retriever. The XOR-TyDi QA dev split is used for generating paired questions across languages and for analysing the different types of inconsistencies observed.

\subsubsection{Evaluation Metric}
We follow the same setup as \citet{lewis2019mlqa}. For Chinese, we use the \texttt{jieba} package to perform tokenization; for all other languages, we use white-space tokenization. We remove punctuation characters by checking the corresponding character type in Unicode. We adopt the mean token F1 score to estimate the performance metrics in line with previous extractive QA work to evaluate multilingual question answering accuracy. We also examine the discrepancies in the models by comparing the percentage of answers that are different. Finally, we manually annotate the types of discrepancies/errors observed.

\subsubsection{Experimental Configuration}
For fine-tuning the multilingual Dense Passage Retriever, we first fine-tune a pretrained multilingual BERT on the Natural Questions (NQ) \cite{NaturalQuestions} dataset. Then we fine-tune our mDPR on the XOR-TyDi QA training split. For fine-tuning the multilingual Answer Generator, we use a pretrained MT5 and then fine-tune it on the TyDi QA training split. 
We use the AdamW optimizer with a learning rate of $3e-5$ for fine-tuning the answer generator, with a batch size of $12$ and two training epochs.
\subsection{Multilingual QA Results}
\label{subsec:result}

We describe our analysis for multilingual question answering results as below. We identify five types of inconsistency errors which we categorize as translation issues, retriever issues, answer extractor issues, inconsistent facts, and different levels of granularity. 

\subsubsection{Oracle Setting}
In the oracle setting, we use the ground truth answer as a filter to select the context and the questions to be fed into the answer extractor for analysis as described in Sec.~\ref{sec:analysis}.

\paragraph{Answer Inconsistency Results:}
There are $146$ questions obtained for analysis from a total of $5K$ questions in the dev set. We observe that $66$ questions yield inconsistent answers across languages. We visualize the first $20$ questions for simplicity as shown in Fig.~\ref{fig:gt_filter_dist}. We observe that given questions with same meaning but in different languages (e.g. sample index $52$ shows question in English (\textit{en}) and Finnish (\textit{fi})), the answers could be different (e.g. the answer asked in \textit{fi} is correct while the answer asked in \textit{en} is wrong). This confirms our hypothesis that QA datasets exhibit information discrepancy across languages.

\paragraph{Inconsistency Error Analysis Results:}
The distribution of the error types is shown in Fig.\ref{fig:gt_filter_dist}. We manually analyze the different inconsistency errors, and find that only three question samples consist of inconsistent facts in different languages.

Some error are due to issues with the answer translation. The current pipeline translates the gold answer to the target language to compute F1 and Exact Match scores. While this is reasonable, answer translations are not consistent all the time. For examples, name translations across different languages (for example, \textit{es} $\rightarrow$ \textit{ja}) can be different.  
\subsubsection{Non-Oracle Setting}
In the non-oracle setting, instead of only selecting the context with the gold answer, we return the top scored context for each language in the total top $1,000$ retrieved passages and use them for generating answers as described in Sec.~\ref{sec:analysis}.

\paragraph{Answer Inconsistency Results:}
We perform the same process as mentioned previously for all the $5K$ samples in the dev set. We observe that almost all questions give inconsistent answers in different languages. We visualize the first 200 questions for simplicity as shown in Fig.~\ref{fig:nogt_filter_dist}.

\paragraph{Inconsistency Error Analysis Results:}
The distribution of the error types is shown in Fig.~\ref{fig:nogt_filter_dist}. We manually analyze the different inconsistency errors. As we expand the search space to the top 1,000 retrieved passages, we find that most of the retrieved context is either unrelated or related but does not contain the answer to the question asked. For context that are related, we find that $4$ of the question samples consist of inconsistent facts in different languages.

One of these samples asks the question "What is the British population?". Answers retrieved in different languages present statistics in different years (some language's Wiki have data for $2010$, some has $2007$) whereas the context in the English Wiki gives a detailed break-down of population in different areas. Given these different contexts, the answer extractor returns different answers.

We also observe bias in the retrieved context. For a question that asks about the governor of the state of Punjab, all the retrieved context, although in different languages, is either related to Afghanistan or Pakistan.

\section{Related Work}
\subsection{Open-domain Question Answering}
Open-Domain QA (OpenQA) has been studied closely with research in Natural Language Processing (NLP), Information Retrieval (IR), and Information Extraction (IE). Traditional OpenQA systems mostly follow a pipeline with three stages, i.e. Question Analysis, Document Retrieval and Answer Extraction \cite{openqa_summary}. 

With the advancement of deep learning techniques, OpenQA systems have seen significant developments in every stage.~\citet{lei2018novel} and ~\citet{xia2018novel} used a CNN-based model and a LSTM-based model in order to develop a question classifier. ~\citet{latent_space_models}, ~\citet{karpukhin2020dense} and ~\citet{retrieve_read} have worked on developing neural retrieval models for searching documents in a latent space. With the development of large-scale datasets like SQuAD \cite{rajpurkar2016squad}, there has been significant progress in the area of neural Machine Reading Comprehension (MRC) methods \cite{wang2017gated, bidirectional_attention_flow, devlin2018bert}. The traditional OpenQA systems have now developed into the "Retriever-Reader" architecture \cite{chen2017reading, reinforced_ranker_reader, multi-step, realm}. Here the Retriever retrieves the relevant documents for a given question, hence behaving like an IR system. The Reader, which is mostly a neural MRC model, tries to infer the answer from the retrieved documents. Based on this "Retriever-Reader" architecture, there has been progress along the lines of re-ranking the retrieved documents before feeding them into the Reader \cite{lee2018ranking, lin-etal-2018-denoising}, using an iterative method to retrieve the relevant documents for a question, \cite{feldman2019multi}, and doing end-to-end training on the OpenQA System \cite{lee2019latent}.

\subsection{Multilingual Question Answering}
A lot of research has been going on in the domain of multilingual question answering. Recently,~\citet{asai2020xor} introduced a new task framework named Cross-lingual Open Retrieval Question Answering (XOR QA), that consists of three new tasks involving cross-lingual document retrieval from multilingual resources.

In order to overcome the issue of data-scarcity in non-English languages, several non-English QA datasets have been created like TyDi QA \cite{clark-etal-2020-tydi}, XOR-TyDi QA ~\cite{asai2020xor} etc. Along similar lines is the dataset MLQA \cite{lewis2019mlqa}, which consists of over $12K$ QA instances in English and $5K$ each in other languages (Arabic, German, Spanish, Hindi, Vietnamese and Simplified Chinese). Each QA instance is parallel between at least $4$ languages on average. MLQA serves as a cross-lingual extension to existing extractive QA datasets, given it's built using a novel alignment context strategy on articles from Wikipedia. Recently, ~\citet{exams} developed a cross-lingual and multilingual dataset for high school examinations. ~\citet{liu-etal-2019-xqa} developed XQA which has a great degree of lexical overlap between reference paragraphs and the questions.

\subsection{Bias in Question Answering}
Although current linguistic machine learning models achieve numerically-high performance on the question answering task, they often lack optimization for reducing implicit biases. There are many works that try to analyze or tackle biases in QA. ~\citet{zhucounterfactual} try to capture and eliminate position bias in large language models in question answering. ~\citet{ahn2021mitigating} developed a metric called Categorical Bias score and attempted to reduce ethnic bias in BERT models for multilingual question answering. Recent attempts to tackle bias try to either automatically generate additional language sources to balance the training data distribution ~\cite{chen2020counterfactual, vision_lang} or to learn a biased model with the source of bias as the input, and then attempt to debias it through logits re-weighting or logits ensembling~\cite{clark2019don, unimodal_biases}.

\section{Conclusion}
In this work, we investigate whether multilingual QA tasks expose information consistencies across different languages. We evaluate to what extent each component of the QA system contributes to inconsistent answers on the XOR-TyDi QA benchmark across nine languages. We observe inconsistent answers given questions with the same content, but in different languages. We analyze the potential reasons that could lead to such inconsistencies including retriever issues, answer extractor issues, different levels of granularity, translation issues, and inconsistent facts, of which retriever issues and answer extractor issues are the most prominent. Our analysis provides new insights into inconsistencies in multilingual question answering, and is one step closer towards a more realistic application of this task.

\begin{table*}[t]
    \footnotesize
    \resizebox{\textwidth}{!} {\begin{tabular}{@{}llc@{}}
    \toprule
    English questions                                                        & Russian questions translated into English                                                    & Cosine similarity \\ \midrule
    How many people died during WW1?                               & How many Germans were captured in the Battle of Stalingrad? & 0.71              \\
    When was the Nazi party founded?                               & When was the KPSS founded?                                  & 0.78              \\
    When was the University of Maryland football team established? & When was the Rome Football Club established?                & 0.78              \\
    When was Siri introduced by Apple?                             & What year was Apple formed?                                 & 0.74              \\
    When was the term anthropology first used?                     & What does the word "anthropology" mean?                     & 0.72              \\ \bottomrule
    \end{tabular}}
    \caption{An example of the question pairs from the English and Russian subsets of TyDi QA found by using question embedding cosine similarity as the metric.}
    \label{tab:cos_sim_eg}
\end{table*}
\begin{figure*}[t]
	\centering
	\begin{subfigure}[]{0.48\linewidth}
        \includegraphics[width=\linewidth]{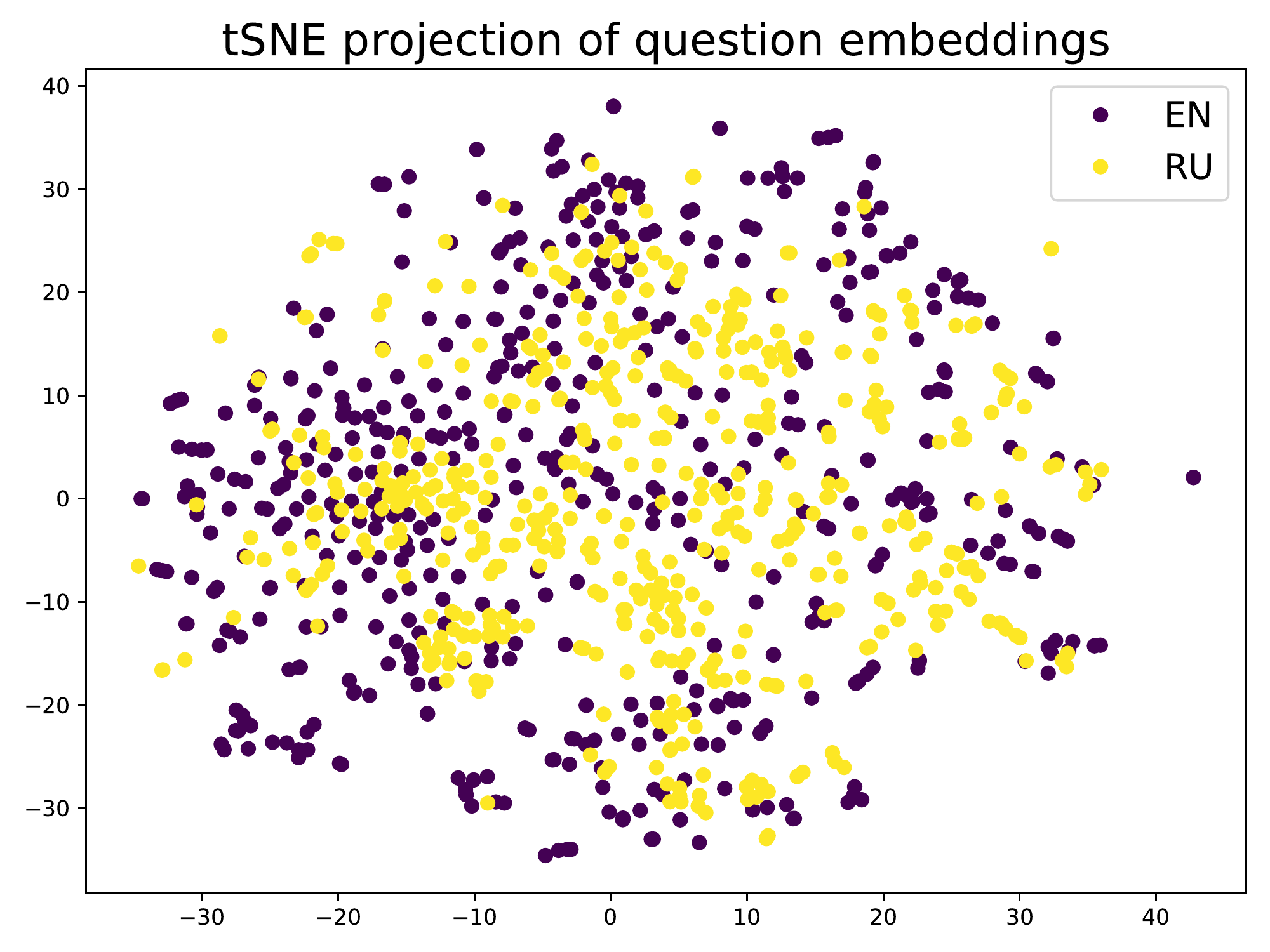}
        \caption{}
        \label{fig:en_ru_tsne}
    \end{subfigure}
	\hfill
	\begin{subfigure}[]{0.48\linewidth}
		\centering
		\includegraphics[width=\linewidth]{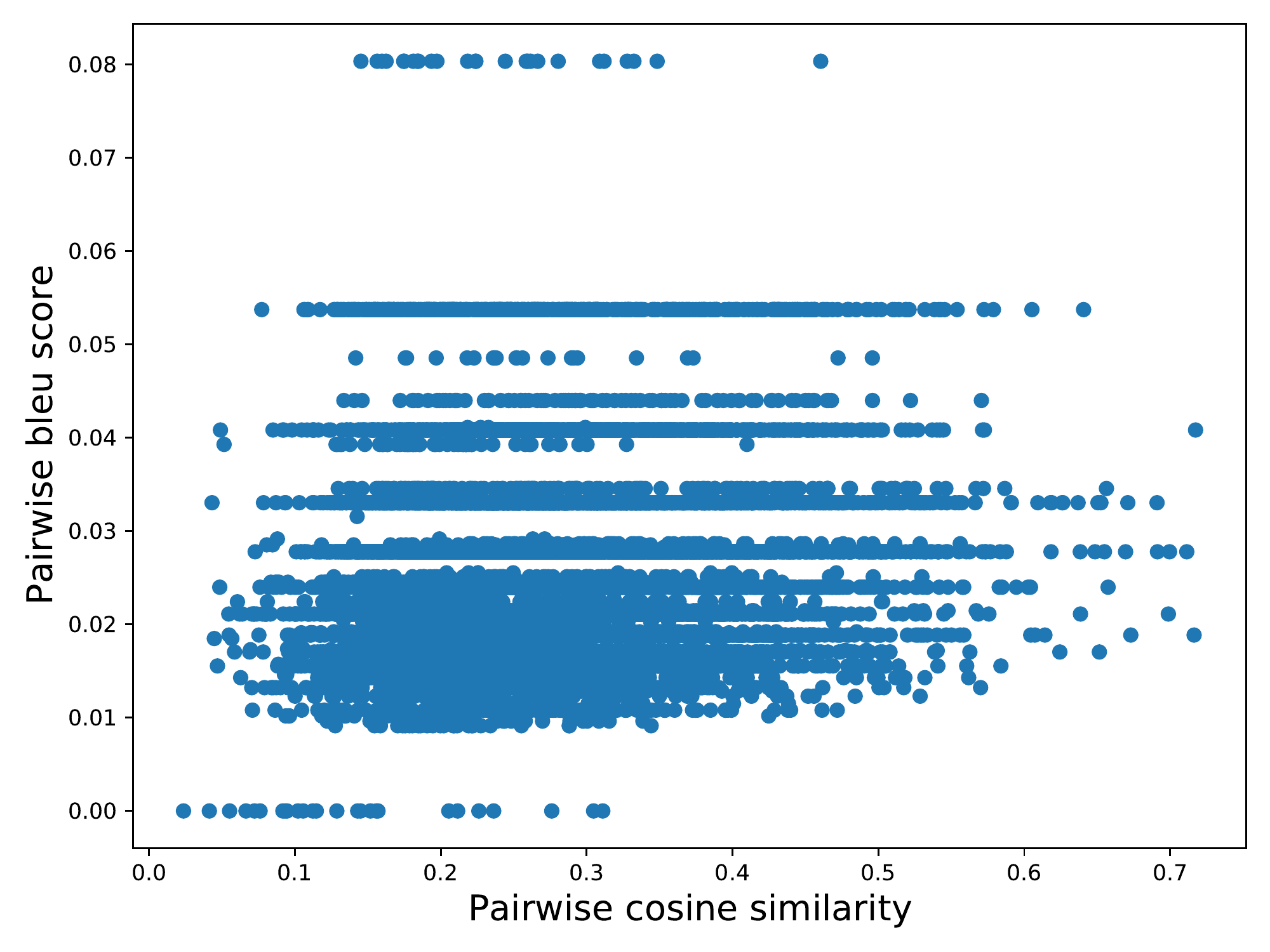}
		\caption{}
		\label{fig:cos_vs_bleu}
	\end{subfigure}
	\caption{(a)t-SNE projection of 500 English question embeddings and 500 Russian question embeddings. The well-mixed scatters indicate the existence of similar questions across English and Russian. (b) No significant correlation between sentence embedding similarity and translation BLEU score is observed.}
	\label{fig:find_sim_qn}
\end{figure*}
\begin{figure*}[t]
	\centering
	\begin{subfigure}[]{0.48\linewidth}
        \includegraphics[width=\linewidth]{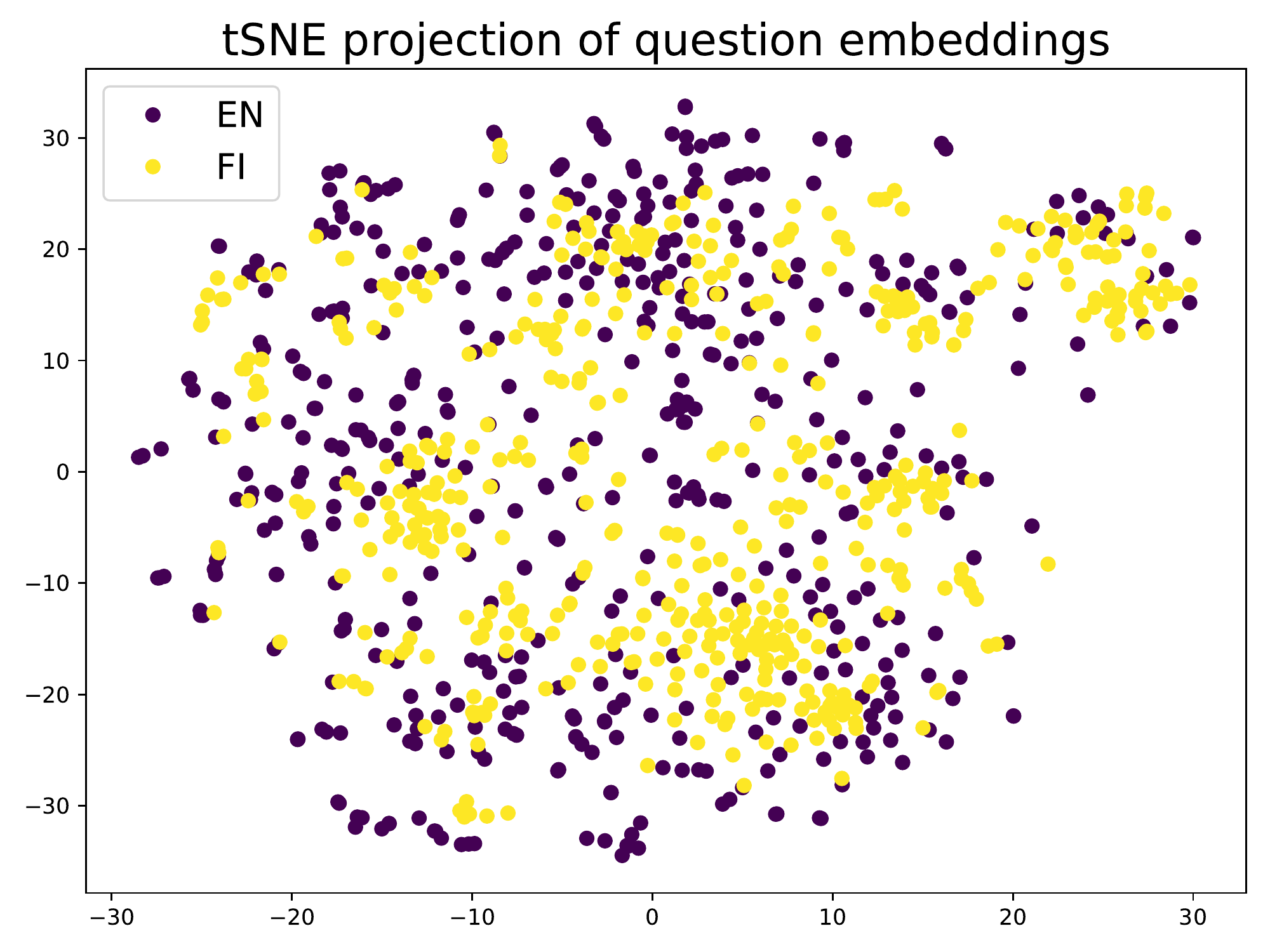}
        \caption{}
        \label{fig:en_ru_tsne}
    \end{subfigure}
	\hfill
	\begin{subfigure}[]{0.48\linewidth}
		\centering
		\includegraphics[width=\linewidth]{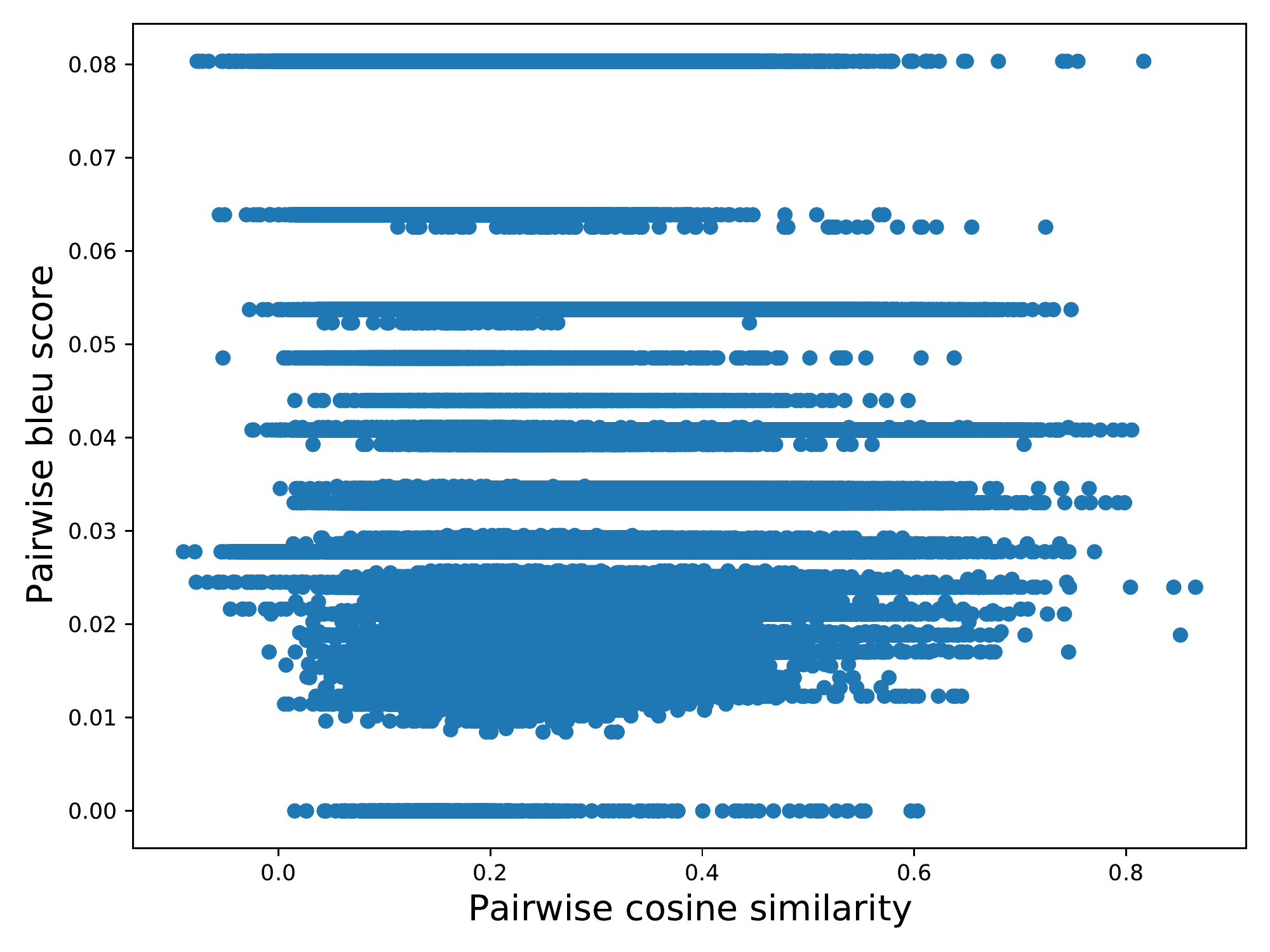}
		\caption{}
		\label{fig:cos_vs_bleu}
	\end{subfigure}
	\caption{(a)t-SNE projection of 500 English question embeddings and 500 Finnish question embeddings. The well-mixed scatters indicate the existence of similar questions across English and Finnish. (b) No significant correlation between sentence embedding similarity and translation BLEU score is observed.}
	\label{fig:find_sim_qn_finish}
\end{figure*}

\bibliography{anthology,custom}
\appendix
\section{Multilingual Question Answering Examples}
\begin{figure*}[t]
	\centering
    \includegraphics[width=\linewidth]{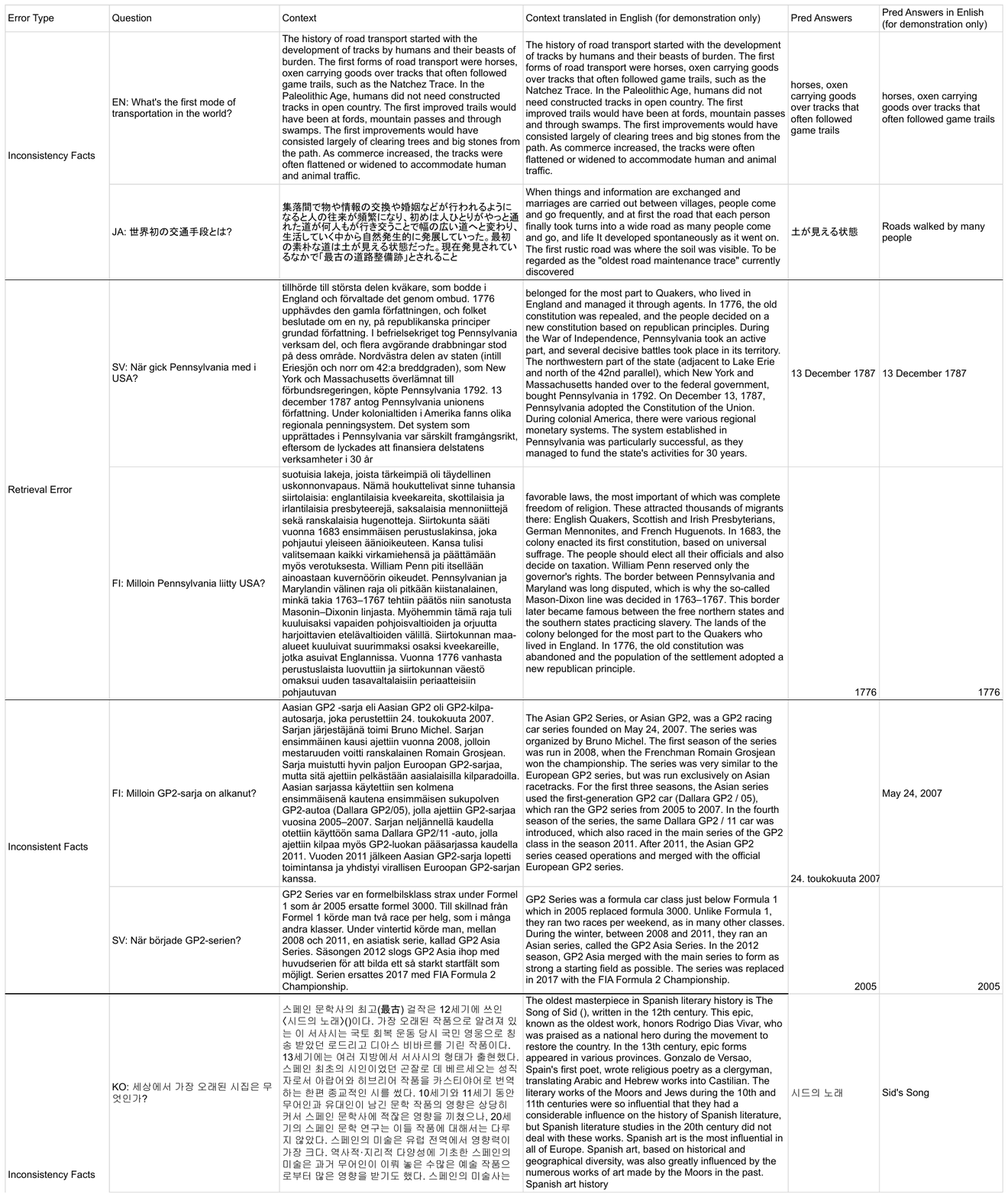}
	\caption{Multilingual Question Answering Examples with Analysis}
	\label{fig:eg}
\end{figure*}

\begin{figure*}[t]
	\centering
    \includegraphics[width=\linewidth]{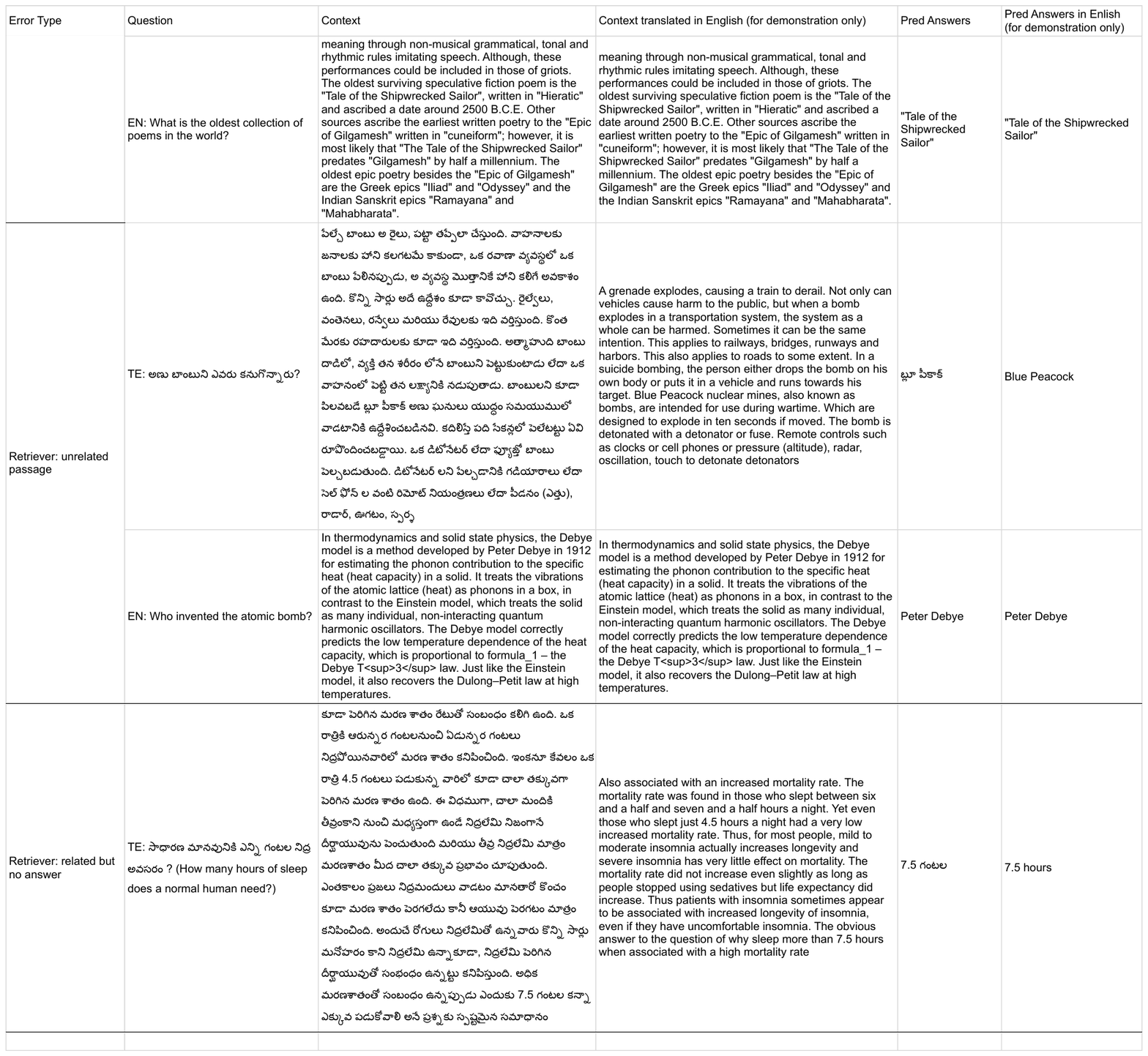}
	\caption{Multilingual Question Answering Examples with Analysis}
	\label{fig:eg2}
\end{figure*}

We provide more multilingual question answering examples with analysis in Fig.~\ref{fig:eg} and Fig.~\ref{fig:eg2}.

\section{Crafting Multilingual QA Test set}
\subsection{Finding similar questions} 
\label{subsec:find_sim_question}
TyDi QA~\cite{clark2020tydi} is a benchmark for information-seeking question answering in typologically diverse languages, including English and Russian.
We first attempt to find a subset of questions that are present in multiple languages but with potentially controversial and different answers. To identify similar questions across different languages, we compare their contextual sentence embeddings~\cite{reimers-2019-sentence-bert}. For each English question $Q^{en}$ and each Russian question $Q^{ru}$, we obtain their respective sentence embeddings $\mathbf{e}^{en}, \mathbf{e}^{ru} \in \mathbb{R}^{768}$. We then compute the pairwise cosine similarities of these embeddings to find similar questions.

Similarly, we project 500 English question embeddings and 500 Russian question embeddings via t-SNE as shown in Fig.~\ref{fig:en_ru_tsne}. Through this visualization, we see that there exist some embeddings that are quite close to each other. The inseparability of the two clusters in this two-dimensional space indicates that some questions share considerable semantic and syntactic similarities.

We then filter the question pairs whose embeddings have cosine similarity greater than 0.7. To evaluate if it is effective to use the sentence embeddings cosine similarity as a metric for question similarity, we translate the Russian questions into English~\cite{TiedemannThottingal:EAMT2020}. We hypothesize that a higher cosine similarity between a pair of English and Russian question embeddings would lead to a higher BLEU score after we translate. However, Fig.~\ref{fig:cos_vs_bleu} does not indicate this trend. There is little correlation between sentence embedding cosine similarity and the BLEU score. This result is likely due to the short length of questions and a small overlap of the same named entities in questions across the two languages. We observe a similar trend between English and Finnish as well as shown in Fig.~\ref{fig:find_sim_qn_finish}.

Nonetheless, using the cosine similarity as a metric gives us some questions on similar topics. We illustrate a few examples in Table.~\ref{tab:cos_sim_eg}. We find that questions about similar topics tend to have higher cosine similarities. For example, both the English and the Russian questions ask about wars in history or sports clubs. Through these limited number of examples, we already see some different cultural divergences in the two subsets. For example, the question "When was the KPSS founded?" is equivalent to "When is the Communist Party of the Soviet Union established?" This question is more culturally specific to Russia and will be less likely to appear in the English question subset.
\end{document}